\documentclass[sigconf]{acmart}

\AtBeginDocument{%
  \providecommand\BibTeX{{%
    \normalfont B\kern-0.5em{\scshape i\kern-0.25em b}\kern-0.8em\TeX}}}

\copyrightyear{2020}
\acmYear{2020}
\setcopyright{acmcopyright}
\acmConference[MM '20]{Proceedings of the 28th ACM International Conference on Multimedia}{October 12--16, 2020}{Seattle, WA, USA}
\acmBooktitle{Proceedings of the 28th ACM International Conference on Multimedia (MM '20), October 12--16, 2020, Seattle, WA, USA}
\acmPrice{15.00}
\acmDOI{10.1145/3394171.3413977}
\acmISBN{978-1-4503-7988-5/20/10}

\usepackage{epsfig}
\usepackage{graphicx}
\usepackage{amsmath}

\usepackage{amssymb}
\usepackage[ruled,vlined]{algorithm2e}
\usepackage{multirow}
\usepackage{makecell}
\usepackage{url}
\usepackage{booktabs}
\usepackage{caption}
\usepackage{subcaption}
\usepackage{libertine}

\begin{document}
\fancyhead{}
\title{Deep Multimodal Neural Architecture Search}


\author{Zhou Yu, Yuhao Cui, Jun Yu}
\authornote{Jun Yu is the corresponding author.}
\affiliation{%
    \institution{Key Laboratory of Complex Systems Modeling and Simulation, School of Computer Science and Technology, Hangzhou Dianzi University, China}
  }
\email{{yuz, cuiyh, yujun}@hdu.edu.cn}



\author{Meng Wang}
\affiliation{%
  \institution{School of Computer Science and Information Engineering, Hefei University of Technology, China}
  }
\email{eric.mengwang@gmail.com}

\author{Dacheng Tao}
\affiliation{%
  \institution{UBTECH Sydney AI Centre, School of Computer Science, FEIT, The University of Sydney, Australia}
  }
\email{dacheng.tao@sydney.edu.au}

\author{Qi Tian}
\affiliation{%
  \institution{Huawei Cloud \& AI}
  \country{China}
  }
\email{tian.qi1@huawei.com}


\begin{abstract}
Designing effective neural networks is fundamentally important in deep multimodal learning. Most existing works focus on a single task and design neural architectures manually, which are highly task-specific and hard to generalize to different tasks. In this paper, we devise a generalized deep multimodal neural architecture search (MMnas) framework for various multimodal learning tasks. Given multimodal input, we first define a set of primitive operations, and then construct a deep encoder-decoder based unified backbone, where each encoder or decoder block corresponds to an operation searched from a predefined operation pool. On top of the unified backbone, we attach task-specific heads to tackle different multimodal learning tasks. By using a gradient-based NAS algorithm, the optimal architectures for different tasks are learned efficiently. Extensive ablation studies, comprehensive analysis, and comparative experimental results show that the obtained MMnasNet significantly outperforms existing state-of-the-art approaches across three multimodal learning tasks (over five datasets), including visual question answering, image-text matching, and visual grounding.
\end{abstract}

\ccsdesc[500]{Computing methodologies~Multi-task learning}
\ccsdesc[500]{Computing methodologies~Neural networks}
\keywords{multimodal learning; neural networks; neural architecture search}


\maketitle

\section{Introduction}
The developments in deep neural networks enable the machine to deal with complicated multimodal learning tasks that require a fine-grained understanding of both vision and language clues, \emph{e.g.}, visual captioning \cite{xu2015show, anderson2017up-down}, visual grounding \cite{rohrbach2016grounding, yu2018mattnet}, image-text matching \cite{kim2018bilinear, nam2017dual}, and visual question answering (VQA) \cite{fukui2016multimodal, yu2017mfb}. Existing approaches have pushed state-of-the-art performance on respective tasks, however, their architectures are usually dedicated to one specific task, preventing them from being generalized to other tasks. This phenomenon raises a question: \emph{Is it possible to design a generalized framework that can simultaneously adapt to various multimodal learning tasks?}

\begin{figure}
\begin{center}
\includegraphics[width=0.45\textwidth]{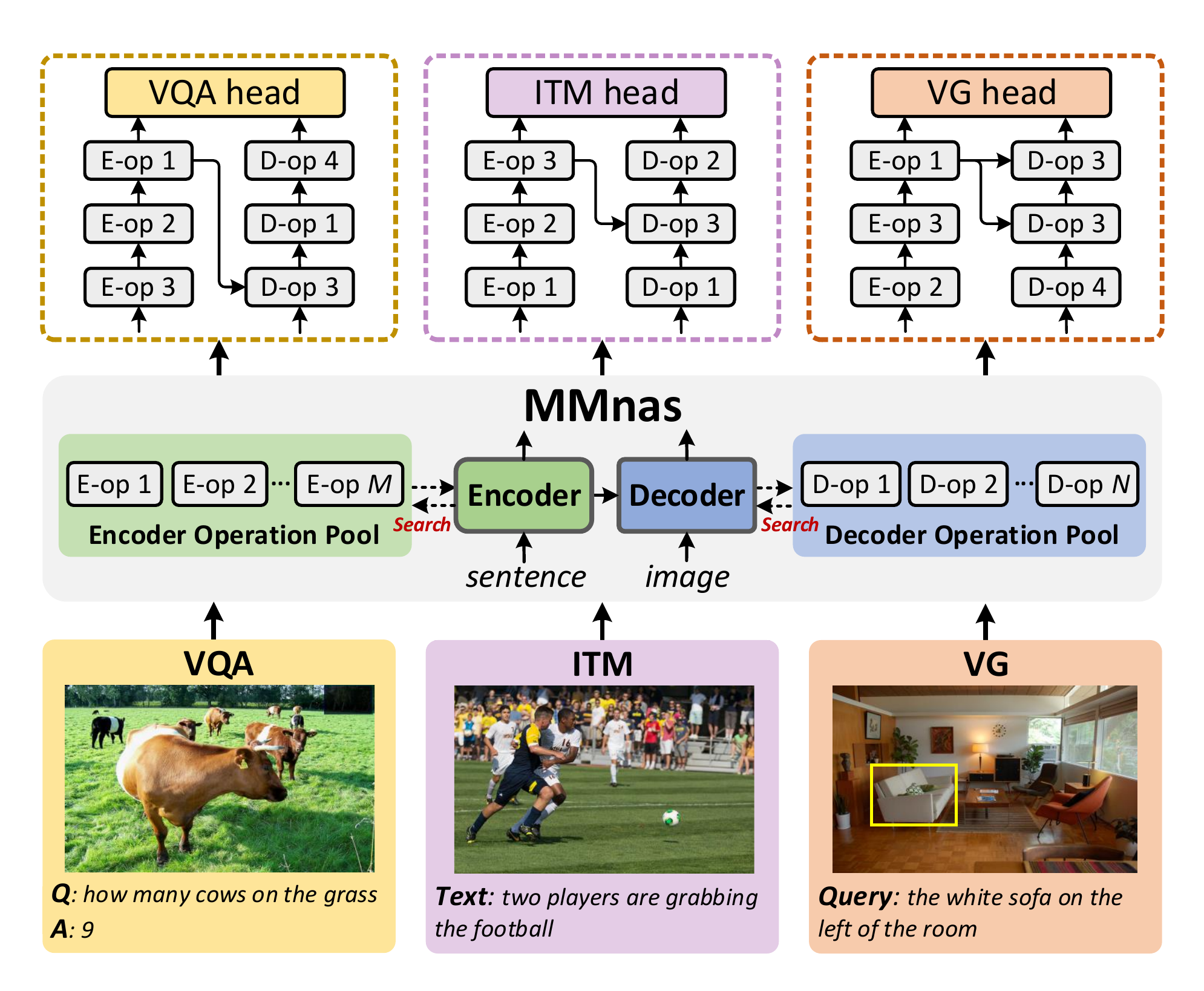}
\caption{Schematic of the proposed generalized MMnas framework, which searches for the optimal architectures for the VQA,  image-text matching, and visual grounding tasks.}
\label{fig:example}
\end{center}
\end{figure}

One promising answer to this question is the \emph{multimodal-BERT} framework \cite{tan2019lxmert, chen2019uniter, lu2019vilbert, li2019visualbert}, which is inspired by the de facto BERT model \cite{devlin2019bert} in the natural language processing (NLP) community. Similar to the transfer learning paradigm \cite{zamir2018taskonomy, tang2016generalized}, BERT adopts a two-stage learning paradigm that first pre-trains a universal backbone via self-supervised learning, and then fine-tune the model for the specific task via supervised learning. Analogously, the multimodal-BERT family pre-trains the Transformer-based backbone to obtain generalizable representations from a large-scale corpus consisting of paired multimodal data (\emph{e.g.}, images and their associated captions). Thereafter, the generalized multimodal backbone is fine-tuned to downstream tasks such as VQA and visual grounding. Despite that the multimodal-BERT approaches deliver promising results on the benchmarks of various multimodal learning tasks, their computational costs are usually very high (\emph{e.g.}, $\sim$10M training samples \cite{tan2019lxmert} or $\sim$300M model size \cite{lu2019vilbert, chen2019uniter}), which severely limits their applicability.

In this paper, we tackle the generalized multimodal learning problem from another perspective. Rather than pre-training one generalized model for various tasks, we design a generalized framework instead, which can adaptively learn the optimal architecture for various tasks. To do this, we introduce neural architecture search (NAS) \cite{zoph2016neural} into multimodal learning and propose a deep multimodal neural architecture search (MMnas) framework (see Figure \ref{fig:example}). Inspired by the modularized MCAN model \cite{yu2019mcan}, we first define a set of primitive operations as the basic unit to be searched. Taking image and sentence features as inputs, we design a unified encoder-decoder backbone by respectively feeding features into the encoder and decoder. The encoder (or the decoder) consists of multiple encoder (or decoder) blocks cascaded in depth, where each block corresponds to an operation searched from the encoder operation pool. On top of the unified backbone, task-specific heads are respectively designed for each task (\emph{e.g.}, VQA, visual grounding). By attaching the unified backbone with each head (\emph{i.e.}, task), we use a gradient-based one-shot NAS algorithm to search the optimal architecture to the respective task. Compared to the \emph{hand-crafted} architecture of MCAN, the \emph{automatically searched} architecture of MMnas can better fit the characteristics of each task and hence lead to better performance. It is worth noting that the proposed MMnas framework is not conflict with the multimodal-BERT approaches. We can also apply the pre-training strategy on the searched architecture to further enhance its performance.

To summarize, the main contributions of this study is three-fold:
\begin{enumerate}
  \item We put forward a new generalized multimodal learning paradigm that uses the neural architecture search (NAS) algorithm to search for the optimal architecture for different tasks. Compared with the multimodal-BERT approaches that use large-scale data to pre-train a generalized model, our paradigm can better capture the characteristics of each task and be more parametric efficient.
  \item We devise a novel MMnas framework, which consists of a unified encoder-decoder backbone and task-specific heads to deal with different task, including visual question answering, image-text matching, and visual grounding.
  \item We conduct extensive experiments on five commonly used benchmark datasets. The optimal MMnasNet delivers new state-of-the-art performance, highlighting the effectiveness and generalizability of the proposed MMnas framework.
\end{enumerate}

\section{Related Work}
We briefly review previous studies on typical multimodal learning tasks and neural architecture search.

\noindent\textbf{Multimodal Learning Tasks:} Multimodal learning aims to build models that can understand and associate information from multiple modalities. From early research on audio-visual speech recognition \cite{yuhas1989integration, dupont2000audio} to the recent explosion of interest in vision-and-language tasks \cite{antol2015vqa, chen2015microsoft, yu2016modeling}, multimodal learning is a multi-disciplinary field of significant importance and potential. At present, multimodal learning with deep neural networks is the de facto paradigm for modern multimodal learning tasks, such as visual question answering (VQA) \cite{antol2015vqa}\cite{kim2018bilinear}\cite{yu2019mcan}, image-text matching \cite{karpathy2015deep, lee2018stacked}, and visual grounding \cite{yu2017joint}\cite{yu2018mattnet}. In the following, we briefly describe three typical multimodal learning tasks and a few representative approaches accordingly.

The VQA task aims to answer a question in natural language with respect to a given image, which requires a fine-grained and simultaneous understanding of both image and question. Antol \emph{et al.} present a large-scale VQA benchmark with human annotations and some baseline methods \cite{antol2015vqa}. Fukui \emph{et al.} \cite{fukui2016multimodal}, Kim \emph{et al.} \cite{kim2016hadamard}, Ben \emph{et al.} \cite{ben2017mutan}, and Yu \emph{et al.} \cite{yu2017mfb} devise different approximated bilinear pooling models to effectively fuse multimodal features with second-order interactions and then integrate them with attention-based neural networks. Most recently, deep co-attention models are proposed to integrate multimodal fusion and attention learning and deliver new state-of-the-art performance on the benchmark datasets \cite{nguyen2018improved,kim2018bilinear,gao2019dynamic}. Yu \emph{et al.} introduce the idea of modularization into the deep co-attention model to characterize the fine-grained multimodal interactions by modularized attention units \cite{yu2019mcan}.

Image-text matching aims to learn two respective mapping functions for the image modality and the text modality, which are then projected into a common semantic space for distance measurement. Karpathy \emph{et al.} propose a deep fragment embedding approach to learn the fine-grained similarity between the visual object in the image and textual word in the caption by maximizing their dot-product similarity under a multi-instance learning framework \cite{karpathy2015deep}. Lee \emph{et al.} propose a stacked cross attention network to exploit the correspondences between textual words and image regions in discovering full latent alignments \cite{lee2018stacked}.

Visual grounding (\emph{a.k.a}, referring expression comprehension) aims to localize an object in an image referred to by a textual query. Rohrbach \emph{et al.} propose a GroundeR model to localize the referred object by reconstructing the sentence using attention mechanism \cite{rohrbach2016grounding}. Yu \emph{et al.} introduce a modular attention network that simultaneously models the language-based attention and visual-based attention to capture rich contextual information for accurate localization \cite{yu2018mattnet}.

The tasks above have the same input modalities (\emph{i.e.}, image and text), however, their solutions are diverse and task-specific, thus preventing them from being generalized to other tasks. Inspired by the success of BERT model \cite{devlin2019bert} in the NLP community, multimodal-BERT approaches are proposed to learn generalized multimodal representation in a self-supervised manner \cite{tan2019lxmert, chen2019uniter, lu2019vilbert, li2019visualbert}. Although promising results have been delivered, these methods usually suffer from tremendous computational costs which limits their usability in practical scenarios.

\captionsetup[subfigure]{font=small}
\begin{figure*}
    \centering
    \begin{subfigure}[h]{0.62\linewidth} 
        \includegraphics[width=\linewidth]{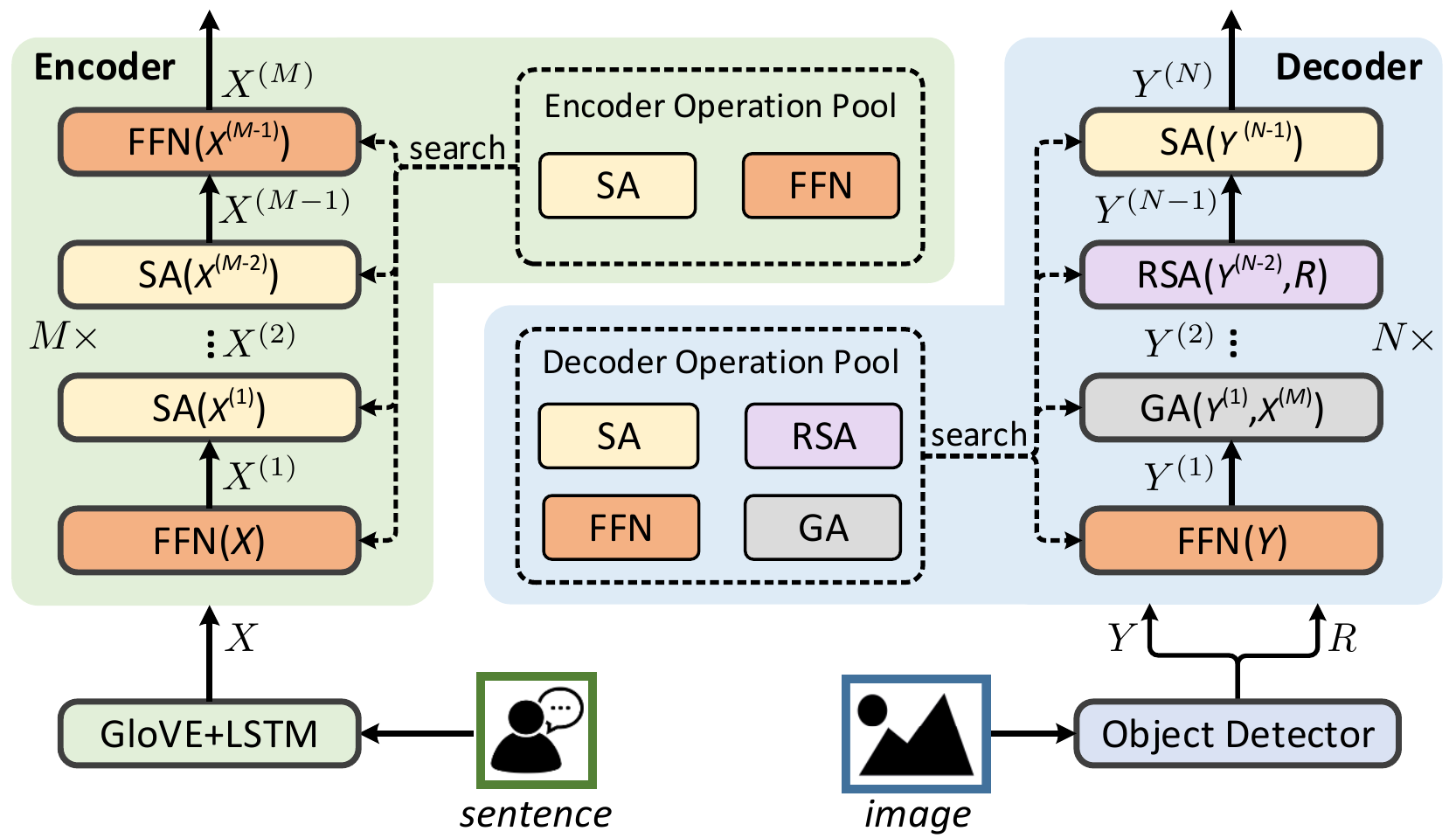}
        \caption{Unified Encoder-Decoder Backbone}\label{fig:backbone}
    \end{subfigure}
        \begin{subfigure}[h]{0.36\linewidth} 
        \includegraphics[width=\linewidth]{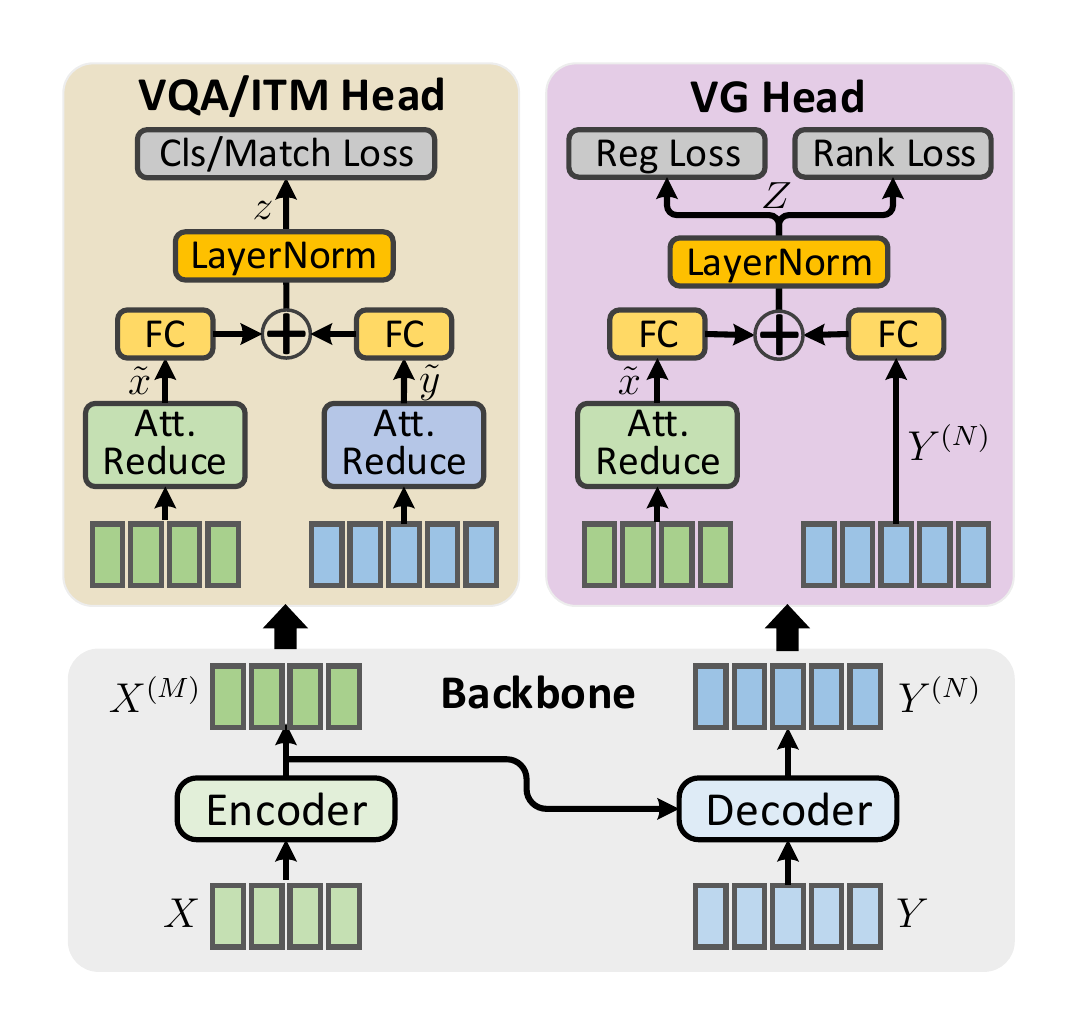}
        \caption{Task-specific Heads}\label{fig:heads}
    \end{subfigure}
    \caption{The flowchart of the MMnas framework, which consists of (a) unified encoder-decoder backbone and (b) task-specific heads on top the backbone for visual question answer (VQA), image-text matching (ITM), and visual grounding (VG).}
    \label{fig:medn}
\end{figure*}

\noindent\textbf{Neural Architecture Search:} Neural architecture search (NAS), \emph{a.k.a.} AutoML, has drawn an increasing interest in the last couple of years, and has been successfully applied to various deep learning tasks, such as image recognition \cite{zoph2018learning}, object detection \cite{ghiasi2019fpn}, and language modeling \cite{so2019evolved}. Early NAS methods use the reinforcement learning to search neural architectures, which are computationally exhaustive \cite{zoph2016neural, zoph2018learning}. Recent works accelerate the searching process by using weight-sharing \cite{pham2018efficient} or hypernetwork \cite{brock2018smash}. Although these methods bring acceleration by orders of magnitude, they usually require a meta-controller (\emph{e.g.}, a hypernetwork or an RNN) which still burdens computational speed. Recently, one-shot NAS methods have been proposed to eliminate the meta-controller by modeling the NAS problem as a single training process of an over-parameterized supernet that comprises all candidate paths \cite{bender2018understanding, liu2018darts, xu2019pc, cai2018proxylessnas}.

The most closely related study to our work is the MFAS approach \cite{perez2019mfas}, which also incorporates NAS to search the optimal architecture for multimodal tasks. However, MFAS focuses on a simpler problem to search for the multimodal fusion model given two input features, which cannot be directly used to address the multimodal learning tasks in this paper.

\section{The MMnas Framework}
In this section, we introduce a generalized multimodal learning framework MMnas via neural architecture search, which can be flexibly adapted to a wide range of multimodal learning tasks involving image-sentence inputs. As shown in Figure \ref{fig:medn}, MMnas contains a unified encoder-decoder backbone and task-specific heads. Taking an image and its associated sentence (\emph{e.g.}, a question, a caption or a query) as inputs, the unified encoder-decoder backbone learns the multimodal interactions with a deep modularized network consisting of stacked encoder and decoder blocks, where each block is searched within a set of predefined primitive operations. On top of the unified backbone, we design task-specific heads to deal with the VQA, image-text matching (ITM), and visual grounding (VG) tasks, respectively. Before presenting the MMnas framework, we first introduce its basic building blocks, the primitive operations.

\subsection{Primitive Operations}
In the following, we present four types of primitive operations, termed as the \emph{self-attention (SA)}, \emph{guided-attention (GA)}, \emph{feed-forward network (FFN)}, and \emph{relation self-attention (RSA)} operations. First, we introduce a {generalized formulation} of the \emph{scaled dot-product attention} proposed in \cite{vaswani2017attention}, which is the core of our primitive operations below.

Denote $m$ queries and $n$ key-value pairs as $Q\in\mathbb{R}^{m\times d}$, $K\in\mathbb{R}^{n\times d}$, $V\in\mathbb{R}^{n\times d}$ respectively, where $d$ is the common dimensionality. The original scaled dot-product attention function in \cite{vaswani2017attention} obtains the output features $F\in\mathbb{R}^{m\times d}$ by weighted summation over all values $V$ with respect to the attention learned from the scaled dot-product of $Q$ and $K$:
\begin{equation}\label{eq:scaled_dot_product}
F = \textsf{A}(Q,K,V)=\mathrm{softmax}(\frac{QK^T}{\sqrt{d}})V
\end{equation}
Inspired by \cite{hu2018relation}, we introduce the apriori relationship $R\in\mathbb{R}^{m\times n}$ between $Q$ and $K$ into Eq.(\ref{eq:scaled_dot_product}) to obtain a more generalized formula:
\begin{equation}\label{eq:sdp_relation}
F = \textsf{A}(Q,K,V,R)=\mathrm{softmax}(\frac{QK^T}{\sqrt{d}}+R)V
\end{equation}

Without loss of generality, the commonly used multi-head mechanism \cite{vaswani2017attention} can also be incorporated with the generalized scaled dot-product attention function, which consists of $h$ paralleled heads (\emph{i.e.}, independent attention functions) to further improve the representation capacity of the attended features:
\begin{equation}\label{eq:ma}
F = \textsf{MHA}(Q,K,V,R)=[\mathrm{head}_1,\mathrm{head}_2,...,\mathrm{head}_h]W^o
\end{equation}
where each $\mathrm{head}_j=\textsf{A}(QW_j^Q,KW_j^K,KW_j^V, R)$ refers to an independent scaled dot-product attention function. $W_j^Q, W_j^K, W_j^V \in\mathbb{R}^{d \times d_h}$ are the projection matrices for the $j$-th head, and $W^o\in\mathbb{R}^{h*d_h \times d}$. $d_h$ is the dimensionality of the output features from each head and is usually set to $d_h=d/h$.

\noindent\textbf{SA(\textit{X}):} Taking a group of input features $X\in\mathbb{R}^{m \times d_x}$ of dimension $d_x$, the output features $Z\in\mathbb{R}^{m \times d}$ of the \textsf{SA} operation are obtained by feeding the inputs through Eq.(\ref{eq:ma}) as follows:
\begin{equation}\label{eq:sa}
Z = \textsf{SA}(X) = \textsf{MHA}(X, X, X, \textbf{0})
\end{equation}
where each $z_i\in Z$ encodes the intra-modal interactions between $x_i$ and all features within $X$. \textbf{0} is an all-zero matrix indicating that no relation prior is provided.

\noindent\textbf{GA(\textit{X, Y}):} Taking two group of features $X\in\mathbb{R}^{m\times d_x}$ and  $Y\in\mathbb{R}^{n\times d_y}$ of dimension $d_x$ and $d_y$ respectively, the \textsf{GA} operation transforms them into $Z\in\mathbb{R}^{m \times d}$ as follows:
\begin{equation}\label{eq:ga}
Z = \textsf{GA}(X, Y) = \textsf{MHA}(X, Y, Y, \textbf{0})
\end{equation}
where each $z_i\in Z$ encodes the inter-modal interactions between $x_i$ and all features within $Y$.

\noindent\textbf{FFN(\textit{X}):} This operation is a two-layer MLP network with ReLU activation and dropout in between. Taking one group of input features $X\in\mathbb{R}^{m \times d_x}$, the transformed output features $Z\in\mathbb{R}^{m \times d}$ of the \textsf{FFN} operation are obtained as follows:
\begin{equation}\label{eq:ffn}
\begin{aligned}
Z= \textsf{FFN}(X) = \mathrm{FC}_{d}\circ\mathrm{Drop}_{0.1}\circ\mathrm{ReLU}\circ\mathrm{FC}_{4d}(X)
\end{aligned}
\end{equation}
where $\mathrm{FC}_{d}(\cdot)$ is a fully-connected layer of output dimension $d$ and $\mathrm{Drop}_{p}(\cdot)$ is a dropout layer with dropout rate $p$. The symbol $\circ$ denotes a composition of two layers.

\noindent\textbf{RSA(\textit{X, R}):} This operation takes a group of features $X\in\mathbb{R}^{m \times d_x}$ along with their relation features $R\in\mathbb{R}^{m\times m\times d_r} $ as inputs, where $d_r$ is the dimensionality of the relation features. The output features $Z\in\mathbb{R}^{m \times d}$ of the \textsf{RSA} operation are obtained as follows:
\begin{equation}\label{eq:rsa}
\begin{aligned}
\textsf{MLP}(R) &=\mathrm{ReLU}\circ\mathrm{FC}_{1}\circ\mathrm{ReLU}\circ\mathrm{FC}_{d_h}(R)\\
Z &= \textsf{RSA}(X, R) = \textsf{MHA}(X, X, X, \textrm{log}(\textsf{MLP}(R)+\epsilon))
\end{aligned}
\end{equation}
where $\textsf{MLP}(R)$ denotes a two-layer MLP network with transformations applied on the last axis of $R$. $\epsilon=1e^{-6}$ is a small constant to avoid the underflow problem.

For each of the primitive operation above, shortcut connection \cite{he2015deep} and layer normalization \cite{ba2016layer} are respectively applied to it. The reason why we use these four operations is two-fold: 1) they are effective and complementary at modeling different kinds of interactions for multimodal learning; and 2) we expect MMnas to be a neat baseline thus only involve the essential operations. Without losing of generality, more effective operations can be included to enlarge the operation pool seamlessly in the future.

\subsection{Unified Encoder-Decoder Backbone}\label{sec:uedb}
Inspired by \cite{yu2019mcan}, we construct a unified encoder-decoder as the backbone to model the deep interactions between the bimodal inputs consisting of an image and its associated sentence. In the following, we describe each component of the backbone in detail.

\noindent\textbf{Sentence and Image Representations:}
The input sentence is first tokenized and then trimmed (or zero-padded) into a sequence of $m$ words. Each word is represented as a 300-D vector using the pre-trained GloVe word embeddings \cite{pennington2014glove}. The word embeddings are fed into a one-layer LSTM network with $d_x$ hidden units, resulting in the final sentence features $X\in\mathbb{R}^{m\times d_x}$.

Following the strategy in \cite{anderson2017up-down}, the input image is represented as a set of objects extracted from a pre-trained object detection model (\emph{e.g.}, Faster R-CNN). For each image, the object detector predicts $n$ objects with each object being represented as a group of visual features and relation features, respectively. The visual features $Y\in\mathbb{R}^{n\times d_y}$ are obtained by pooling the convolutional features from the detected objects. The relation features $R\in\mathbb{R}^{n\times n\times 4}$ are calculated by the relative spatial relationships of object pairs\footnote{Denote the location of the $i$-th object as $\{x_i, y_i, h_i, w_i\}$, where $x_i, y_i$ refer to the center of the object, and $w_i, h_i$ refer to the width and height of the object, respectively. Following the strategy in \cite{hu2018relation}, the 4-D relation feature between the $m$-th object and the $n$-th object is defined as $\left[\mathrm{log}({|x_m-x_n|}/{w_m}), \mathrm{log}({|y_m-y_n|}/{h_m}), \mathrm{log}({w_n}/{w_m}),\mathrm{log}({h_n}/{h_m}) \right]$.}.

\noindent\textbf{Sentence Encoder and Image Decoder:}
Taking the word-level sentence features $X$ as inputs, the sentence encoder learns the intra-modal interactions of sentence words by passing $X$ through $M$ encoder blocks $\{b_\textrm{enc}^{(1)}, b_\textrm{enc}^{(2)}, ..., b_\textrm{enc}^{(M)}\}$ recursively:
\begin{equation}\label{eq:enc}
X^{(i)} = b_\textrm{enc}^{(i)}(X^{(i-1)})
\end{equation}
where $i\in\{1,2,..., M\}$ and $X^{(0)}=X$. Each $b_\textrm{enc}^{(i)}(\cdot)$ corresponds to an operation searched from an encoder operation pool with independent operation weights. Similar to \cite{yu2019mcan}, the encoder operation pool consists of two candidate operations: \textsf{SA} and \textsf{FFN}.

Analogous to the sentence encoder, we design an image decoder consisting of $N$ decoder blocks $\{b_\textrm{dec}^{(1)}, b_\textrm{dec}^{(2)}, ..., b_\textrm{dec}^{(N)}\}$. Slightly different from that of the encoder, the decoder operation pool contains four operations: \textsf{SA}, \textsf{RSA}, \textsf{GA}, and \textsf{FFN}. Taking the visual features $Y$ and relation features $R$ from the image, along with the output features $X^{(M)}$ from the sentence encoder as inputs, the image decoder models the intra- and inter-modal interactions of the multimodal inputs in a recursive manner:
\begin{equation}\label{eq:dec}
Y^{(i)} = b_\textrm{enc}^{(i)}(Y^{(i-1)}, R, X^{(M)})
\end{equation}
where $i\in\{1,2,..., N\}$ and $Y^{(0)}=Y$. Each $b_\textrm{dec}^{(i)}(\cdot)$ takes at least one input (\emph{i.e.}, $Y^{(i-1)}$) and may have an additional input (\emph{i.e.}, $R$ or $X^{(M)}$) if specific operation is searched (\emph{i.e.}, \textsf{RSA} or \textsf{GA}).

\subsection{Task-specific Heads}
The output sentence features $X^{(M)}=[x_1^{(M)}, x_2^{(M)}, ...,x_m^{(M)}]$ and image features $Y^{(N)}=[y_1^{(N)}, y_2^{(N)}, ...,y_n^{(N)}]$ from the unified encoder-decoder backbone contain rich information about the attentive interactions between the sentence words and image objects. On top of the backbone, we attach task-specific heads to address the visual question answering (VQA), image-text matching (ITM), and visual grounding (VG) tasks, respectively.

\noindent\textbf{VQA Head:} Similar to most existing works \cite{antol2015vqa,yu2017mfb,kim2018bilinear}, we resolve the VQA problem by predicting the best-matched answer to the question from a large answer vocabulary. Inspired by the multimodal fusion model in \cite{yu2019mcan}, we use two independent attentional reduction models for $X^{(M)}$ and $Y^{(N)}$ to obtain their reduced features $\tilde{x}$ and $\tilde{y}$, respectively:
\begin{equation}\label{eq:self_att}
\begin{aligned}
\alpha &= \mathrm{softmax}(\textsf{MLP}(X^{(M)})),~~~&\beta&= \mathrm{softmax}(\textsf{MLP}(Y^{(N)})) \\
\tilde{x}&=\sum_{i=1}^m \alpha_ix_i^{(M)}, &\tilde{y}&=\sum_{i=1}^n \beta_iy_i^{(N)}
\end{aligned}
\end{equation}
where $\alpha\in\mathbb{R}^m, \beta\in\mathbb{R}^n$ are the attention weights to be learnt. $\textsf{MLP}(X)=\mathrm{FC}_{1}\circ\mathrm{ReLU}\circ\mathrm{FC}_{d}(X)$ corresponds to a two-layer MLP network. After that, the reduced features are fused together as follows:
\begin{equation}\label{eq:fusion_vqa}
z = \mathrm{LayerNorm}(W_x^T\tilde{x} + W_y^T\tilde{y})
\end{equation}
where $W_x, W_y\in\mathbb{R}^{d \times d_z}$ are two projection matrices to embed the input features into a $d_z$-dimensional common space. LayerNorm is appended on the fused feature to stabilize training \cite{ba2016layer}.

The fused feature $z$ is then projected into a vector $p\in\mathbb{R}^{k}$ and then fed into a $k$-way classification loss, where $k$ denotes the size of the answer vocabulary. For the dataset that provides multiple answers to each question, we formulate it as a multi-label classification problem and use binary cross-entropy (BCE) loss to train the model. For the dataset that only has one answer to each question, we regard it as a single-label classification problem and use the softmax cross-entropy loss instead.

\noindent\textbf{ITM Head:} Image-text matching aims to learn a matching score to measure the cross-modal similarity between the image-text pair. Since the outputs of the ITM and VQA tasks are similar, we therefore reuse part of the model in the VQA head. On top of the fused feature $z$ from Eq.(\ref{eq:fusion_vqa}), the matching score $s\in(0,1)$ is obtained as follows:
\begin{equation}\label{eq:fusion_itm}
s = \sigma(W_z^Tz)
\end{equation}
where $W_z\in\mathbb{R}^{d_z}$ and $\sigma(\cdot)$ denotes the sigmoid function. Denote the predicted matching score of an input image-text pair as $s(\mathcal{I},\mathcal{T})$, where $(\mathcal{I},\mathcal{T})$ represents a positive sample with correspondence. We use BCE loss with hard negatives mining for $(\mathcal{I},\mathcal{T})$ as our loss function to train the matching model:


\begin{equation}\label{eq:loss_itmhead}
\begin{aligned}
 \mathcal{L}_\mathrm{match}&= \mathrm{log}(s(\mathcal{I},\mathcal{T})) +\mathrm{log}(1-s(\mathcal{I},\mathcal{T'}))\\
&+\mathrm{log}(s(\mathcal{I},\mathcal{T})) +\mathrm{log}(1-s(\mathcal{I'},\mathcal{T}))
\end{aligned}
\end{equation}
where $\mathcal{T'}$ and $\mathcal{I'}$ denote the hard negative text and image samples for $(\mathcal{I},\mathcal{T})$ mined from the whole training set per training epoch.

\noindent\textbf{VG Head:} We address the visual grounding task by predicting a ranking score and a refined bounding box for each visual object in the image referred to by the query. To do this, we first feed the word-level query features $X^{(M)}$ into the attentional reduction model in Eq.(\ref{eq:self_att}) to obtain the reduced feature vector $\tilde{x}$. After that, $\tilde{x}$ is broadcasted and integrated with the object-level image features $Y^{(N)}$ as follows:
\begin{equation}\label{eq:vg_fusion}
Z = \mathrm{LayerNorm}(W_x^T\tilde{x} + W_y^TY^{(N)})
\end{equation}
where $Z\in\mathbb{R}^{n \times d_z}$ correspond to the fused features of $n$ objects in the image. Each object feature $z\in Z$ is then linearly projected into a ranking score $s\in\mathbb{R}$ and a 4-D bounding box offset $b\in\mathbb{R}^4$, respectively. Similar to \cite{yu2018rethinking}, we design a multi-task loss function consisting of a ranking loss $\mathcal{L}_\mathrm{rank}$ and a regression loss $\mathcal{L}_\mathrm{reg}$:
\begin{equation}\label{eq:loss_vghead}
\mathcal{L} = \mathcal{L}_\mathrm{rank} + \lambda\mathcal{L}_\mathrm{reg}
\end{equation}
where $\lambda$ is a hyper-parameter to balance the two terms.
The $\mathcal{L}_\mathrm{rank}$ term penalizes the KL-divergence between the predicted scores $S=[s_1, s_2,...,s_n]\in\mathbb{R}^n$ and the ground-truth scores $S^*\in\mathbb{R}^n$ for $n$ objects, where $S^*$ are obtained by calculating the IoU scores of all objects with respect to the unique ground-truth bounding box. Softmax normalizations are respectively applied to $S$ and $S^*$ to form a score distribution.
The $\mathcal{L}_\mathrm{reg}$ term penalizes the smoothed $L_1$ distance \cite{girshick2015fast} between the predicted offset $b$ and the ground-truth offset $b^*$ for the objects with their IoU scores $S^*$ larger than a threshold $\sigma$. The offset $b^*\in\mathbb{R}^4$ is obtained by calculating the translations between the bounding box of the input object and the bounding box of ground-truth object \cite{girshick2015fast}.

\begin{algorithm}
\footnotesize
\SetAlgoLined
\KwIn{A supernet parameterized by the architecture weights $\theta$ and the model weights $W$. Training set $\mathcal{D}_a$ and $\mathcal{D}_m$ are used to optimize $\theta$ and $W$, respectively. $T_w$ and $T_j$ denote the number of epochs for the warming-up and iterative optimization stages, respectively. $u$ is a factor to balance the update frequencies of $\theta$ and $W$.}
\KwOut{The searched optimal architecture $a^*$ }
 Random initialize $\theta$ and $W$\;
\# \emph{The warming-up stage}\;
\For{$t=1$ \KwTo $T_{w}$}{
   Random sample an architecture $a\sim\mathcal{A}$\;
   Random sample a mini-batch $d_{m}\subseteq \mathcal{D}_m$\;
   Update $W$ by descending $\nabla_W\mathcal{L}_\mathrm{train}(\mathcal{N}(a,W))$ on $d_m$\;
}
\# \emph{The iterative optimization stage}\;
\For{$t=1$ \KwTo $T_{j}$}{
    \For{$i=1$ \KwTo $u$}{
        Random sample a mini-batch $d_{m}\subseteq \mathcal{D}_m$\;
        Sample an architecture $a\sim\mathcal{A}(\theta)$ with respect to $\theta$\;
        Update $W$ by descending $\nabla_{W}\mathcal{L}_\mathrm{train}(\mathcal{N}(a,W))$ on $d_m$\;
   }
   Random sample a mini-batch $d_{a}\subseteq \mathcal{D}_a$\;
   Sample an architecture $a\sim\mathcal{A}(\theta)$ with respect to $\theta$\;
   Update $\theta$ by descending $\nabla_\theta\mathcal{L}_\mathrm{train}(\mathcal{N}(a,W))$ on $d_a$\;
}
{Return} $a^*$ by picking the operation with the largest value in $\theta^*$  for each block.
 \caption{Search Algorithm for MMnasNet.}\label{alg:1}
\end{algorithm}

\section{Search Algorithm}\label{sec:search}
To obtain the optimal MMnasNet architecture for each task on specific dataset, we introduce an efficient one-shot search algorithm that search the optimal architecture within an over-parameterized supernet with weight sharing.

Denote a supernet as $\mathcal{N}(\mathcal{A}(\theta), W)$ that encodes the whole search space $\mathcal{A}$ of MMnas, where $W$ and $\theta$ correspond to the model weights and architecture weights of all the possible operations in the supernet, respectively\footnote{Given a MMnas supernet consisting of $M$ encoder blocks and $N $ decoder blocks,  the size of the search space is $2^M$$\times$$4^N$ and the number of all the possible operations in the supernet is $2M$+$4N$, where 2 and 4 correspond to the sizes of the encoder and decoder operation pools, respectively.}. The optimal architecture is obtained by minimizing the expectation with respect to $\theta$ and $W$ jointly:
\begin{equation}\label{eq:nas1}
(\theta^*, W^*) = \mathop{\mathrm{argmin}}\limits_{\theta, W} ~\mathbb{E}_{a\sim\mathcal{A}(\theta)}[\mathcal{L}{\left(\mathcal{N}(a, W)\right)]}
\end{equation}
where for each of the three tasks above, $\mathcal{L}(\mathcal{N}(a,W))$ indicates using the task-specific loss function to optimize the weights of network architecture $\mathcal{N}(a,W)$, where $a$ is a valid architecture sampled from the supernet with respect to $\theta$. Based on the obtained optimal $\theta^*$, the optimal architecture $a^*$ is obtained by selecting the operation with the largest architecture weight in each block of the backbone.

Inspired by the strategy in \cite{cai2018proxylessnas}, we adopt an iterative algorithm to optimize the architecture weights $\theta$ and the model weights $W$ alternatively. We first separate the training set into two non-overlapping sets $\mathcal{D}_m$ and $\mathcal{D}_a$. When training the model weights $W$, we first freeze the architecture weights $\theta$ and stochastically sample exactly one operation for each block with respect to $\theta$ after softmax activation, which results in a valid architecture $a$. After that, we update the model weights $W$ activated by $a$ via standard gradient descent on $\mathcal{D}_m$. When training the architecture weights $\theta$, we freeze the model weights $W$, sample a valid architecture $a$, and then update $\theta$ via gradient descent on $\mathcal{D}_a$.

As claimed in \cite{chu2019fairnas}, the iterative optimization of $W$ and $\theta$ inevitably introduces bias to certain architectures and leave the rest ones poorly optimized. To alleviate the problem, we introduce an additional \emph{warming-up} stage before the iterative optimization. In the warming-up stage, we do not train the architecture weights $\theta$ and sample operations uniformly to train the model weights $W$. This ensures that the model weights $W$ are well initialized thus leading to more impartial and robust architecture search.

The detailed search algorithm is illustrated in Algorithm \ref{alg:1}.

\section{Experiments}
We evaluate the searched MMnasNets on three multimodal learning tasks and perform comparative analysis to the state-of-the-art methods on five benchmark datasets thoroughly. Furthermore, we conduct comprehensive ablation experiments to explore the reasons why MMnas is effective.

\subsection{Datasets}
\noindent\textbf{VQA-v2} is a commonly-used dataset for the VQA task \cite{goyal2016making}. It contains human annotated question-answer pairs for COCO images \cite{lin2014microsoft}. The dataset is split into three subsets: {train} (80k images with 444k questions); {val} (40k images with 214k questions); and {test} (80k images with 448k questions). The {test} subset is further split into {test-dev} and {test-std} sets that are evaluated online. The results consist of three per-type accuracies (Yes/No, Number, and Other) and an overall accuracy.

\noindent\textbf{Flickr30K} contains 31,000 images collected from Flickr website with five captions each. Following the partition strategy of \cite{lee2018stacked,karpathy2015deep}, we use 1,000 images for validation and 1,000 images for testing and the rest for training.

\noindent\textbf{RefCOCO, RefCOCO+ and RefCOCOg} are three datasets to evaluate visual grounding performance. All three datasets are collected from COCO images \cite{lin2014microsoft}. RefCOCO and RefCOCO+ are split into four subsets: train (120k queries), val (11k queries), testA (6k queries about people), and testB (5k queries about objects). RefCOCOg is split into three subsets: train (81k queries), val (5k queries), and test (10k queries)

The statistics and evaluation metrics of the datasets are summarized in Table \ref{table:dts}.

\subsection{Experimental Setup}
\noindent\textbf{Universal Setup:} We use the following hyper-parameters for MMnasNet as the default settings unless otherwise stated. For each primitive operation, the latent dimensionality in the multi-head attention $d$ is 512 and the number of heads $h$ is 8. The dimensionality of the fused features $d_z$ is set to $2d$. The number of encoder blocks $M$ and decoder blocks $N$ are respectively set to 12 and 18 to match the number of blocks in the 6-layer MCAN model\footnote{A $L$-layer MCAN model corresponds to a special case of the MMnasNet model consisting of $2L$ encoder blocks (with repeated \textsf{SA}-\textsf{FFN} operations) and $3L$ decoder blocks (with repeated \textsf{SA}-\textsf{GA}-\textsf{FFN} operations).} \cite{yu2019mcan}.

For each dataset, we use its train split to perform architecture search. The train set is further random split into two subsets $D_m$ and $D_a$ with $|D_m|/|D_a|=4$. Each randomly initialized model is warmed-up for $T_w=50$ epochs and then searched for another $T_j=20$ epochs with the early stopping strategy.
The frequency ratio $u$ for updating the model and architecture weights is set to 5. With the searched optimal architecture, we train the MMnasNet model again from scratch to obtain the final model.

\noindent\textbf{VQA Setup:} For VQA-v2, we follow the setting in \cite{yu2019mcan} that all questions are processed to a maximum length of $m=14$ and the size of the answer vocabulary is set to 3129.  The visual features and relation features are extracted from a pre-trained Faster R-CNN model on Visual Genome \cite{anderson2017up-down}. The number of extracted objects $n\in[10,100]$ is determined by a confidence threshold.

\noindent\textbf{ITM Setup:} For Flickr30K, the maximum length of texts (\emph{i.e.}, captions) is set to $m=50$. The visual features and relation features are extracted from a Faster R-CNN model pre-trained on Visual Genome with the number of objects $n=36$ \cite{anderson2017up-down}. For each positive image-text pair $(\mathcal{I}, \mathcal{T})$ in the training set, we use the following hard sample mining strategy before each training epoch: we randomly sample 64 negative images per text and 64 negative texts per image from the whole training set to generate negative image-text pairs. Thereafter, we feed all these negative pairs to the current model checkpoint to predict their matching scores and regard the top-5 ranked negative samples as the hard negative samples according to their scores. Finally, we randomly pick one hard image sample $\mathcal{I'}$ and one hard text sample $\mathcal{T'}$ from the candidate hard negative samples, respectively.

\noindent\textbf{VG Setup:} We use the same settings for the three visual grounding datasets. For the textual queries, the maximum length is set to $m=14$. For the images, we adopt two pre-trained object detectors to extract the visual features: 1) a Mask R-CNN model trained on COCO \cite{he2017mask}; and 2) a Faster R-CNN model trained on Visual Genome \cite{ren2015faster}. During the training data preparation for the two detectors, we excluded all the images that exist in the training, validation and testing sets of RefCOCO, RefCOCO+, and RefCOCOg to avoid the data leakage problem. For both detectors above, we detect $n=100$ objects for each image to extract the visual and relation features. The loss weight $\lambda$ is set to 1.

\begin{table}
\centering
\footnotesize
\caption{The detailed statistics and evaluation metrics of the tasks and datasets.}\label{table:dts}
 \vspace{-5pt}
\begin{tabular}{c|c|cccc}
\toprule
Task &Dataset& Image Source& \#Img. & \#Sent. & Metric\\
\midrule
{VQA} &VQA-v2 \cite{goyal2016making}& COCO & 204K &1.1M & {Accuracy}\\
\midrule
ITM & Flickr30K \cite{plummer2015flickr30k}& Flickr & 31K & 155K & Recall@K\\
\midrule
\multirow{3}{*}{VG} & RefCOCO \cite{kazemzadeh2014referitgame} & \multirow{3}{*}{COCO} & 20K & 142K & \multirow{3}{*}{Accuracy}\\
& RefCOCO+ \cite{kazemzadeh2014referitgame}&&20K & 142K & \\
& RefCOCOg \cite{mao2016generation}&&26K & 95K & \\
\bottomrule
\end{tabular}
\vspace{-5pt}
\end{table}

\captionsetup[subtable]{font=footnotesize}
\begin{table*}
   \small
\begin{subtable}[t]{.36\textwidth}
		\centering
		\begin{tabular}{c|cccc}
            \toprule
            Decoder operations & All & Y/N& Num&Other\\
            \midrule
            $\{\mathsf{GA}, \mathsf{FFN}\}$ & 66.5 & 84.9 & 45.2 & 58.2\\
           $\{\mathsf{SA}, \mathsf{GA}, \mathsf{FFN}\}$ & 67.4 & 85.0 & 49.7 & 58.7\\
           $\{\mathsf{RSA}, \mathsf{GA}, \mathsf{FFN}\}$ & 67.6 & 84.9& 51.3 & 58.8\\
           $\{\mathsf{SA}, \mathsf{RSA}, \mathsf{GA}, \mathsf{FFN}\}$ & \textbf{67.8} & \textbf{85.1} &\textbf{52.1}& \textbf{58.9} \\
            \bottomrule
        \end{tabular}
        \vspace{5pt}
        \subcaption{\emph{Search Space:} Per-type accuracies of MMnasNet with different decoder operation pools. All models use the same encoder operation pool of $\{\mathsf{SA}, \mathsf{FFN}\}$.}
		\label{table:space}
	\end{subtable}
    \quad
    \quad
    \small
	    \begin{subtable}[t]{.3\textwidth}
		\centering
        \begin{tabular}{cc|cc}
            \toprule
            $M$ & $N$ & MCAN (Size) & MMnasNet (Size)  \\
             \midrule
            4 & 6 & 66.1 (27M) &  67.1 (28M)\\
            8 & 12 & 66.9 (41M) &  67.7 (44M)\\
            12 &18 & 67.2 (56M)& \textbf{67.8} (58M) \\
            16 & 24 & 67.2 (68M) & 67.7 (76M) \\
            \bottomrule
        \end{tabular}
        \vspace{5pt}
        \subcaption{\emph{Model Depth:} Overall accuracies and sizes of MCAN and MMnasNet with different number of encoder blocks $M$ and decoder blocks $N$.}
    \label{table:depth}
	\end{subtable}
\quad
\quad
    \small
	\begin{subtable}[t]{.265\textwidth}
		\centering
        \begin{tabular}{cc|c}
            \toprule
            Encoder & Decoder & Accuracy \\
             \midrule
            R & R & 66.9 \\
            S & R &  67.1\\
            R & S & 67.6 \\
            S & S & \textbf{67.8} \\
            \bottomrule
        \end{tabular}
        \vspace{5pt}
		\subcaption{\emph{Random v.s. Searched:} Overall accuracies of MMnasNet with random (R) or searched (S) architecture for the encoder-decoder.}
        \label{table:random}
	\end{subtable}
	\caption{Ablation experiments for MMnasNet on VQA-v2. We train on the \emph{train} split and report the results on the \emph{val} split.}
     \vspace{-5pt}
    \label{table:aba}
\end{table*}

\subsection{Ablation Experiments}
We run a number of ablations experiments on VQA-v2 to analyze the reason behind MMnasNet's effectiveness. Results shown in Table \ref{table:aba} are discussed in detail next.

\noindent\textbf{Search Space:} In Table \ref{table:space}, we compare the MMnasNet models searched from different decoder operation pools. From the results, we can see that: 1) modeling the intra-modal attention among visual objects by $\textsf{SA}$ or $\textsf{RSA}$ is vital to object counting performance (\emph{i.e.}, the \emph{number} type answers), which is consistent with the results reported in \cite{yu2019mcan}; 2) introducing the $\textsf{RSA}$ operation which models the relative spatial relationships between paired objects can further facilitate the object counting performance; and 3) $\textsf{SA}$ and $\textsf{RSA}$ are complementary to each other, hence modeling them together leads to the best performance on all answer types.

\noindent\textbf{Model Depth:} In Table \ref{table:depth}, we compare MMnasNet to the reference MCAN model \cite{yu2019mcan} under different model depths (\emph{i.e.}, number of encoder blocks $M$ and decoder blocks $N$). The results reveal that: 1) MMnasNet consistently outperforms MCAN, especially when the model depth is relatively shallow (\emph{e.g.}, $M\leq8$). This can be explained that the optimal architectures for different model depths are quite different; 2) with the same $M$ and $N$, the model size of MMnasNet is slightly larger than MCAN. This is because MMnasNet tends to use more \textsf{FFN} operations, which introduces more parameters to increase the nonlinearity of the model; and 3) with the increase of model depth, both MCAN and MMnasNet saturate at $M$=12 and $N$=18, which reflects the bottleneck of the used deep encoder-decoder framework.

\noindent\textbf{Random \textit{vs.} Searched:} To prove the necessity and superiority of the searched architectures over randomly generated ones, we conduct the experiments in Table \ref{table:random} by alternatively using the searched or random architectures for the encoder and decoder, respectively. From the results, we can see that: 1) the searched architectures outperforms the random counterparts by up to 0.9 points; 2) the design of the decoder architecture is much more important than the encoder architecture; and 3) the all-random architecture also performs well compared to some recent works \cite{kim2018bilinear, gao2019dynamic}. This suggests the used primitive operations that constitute the architecture also play a key role in model performance.

\captionsetup[subfigure]{font=small}
\begin{figure}
    \centering
    \begin{subfigure}[h]{0.325\columnwidth}
        \includegraphics[width=\linewidth]{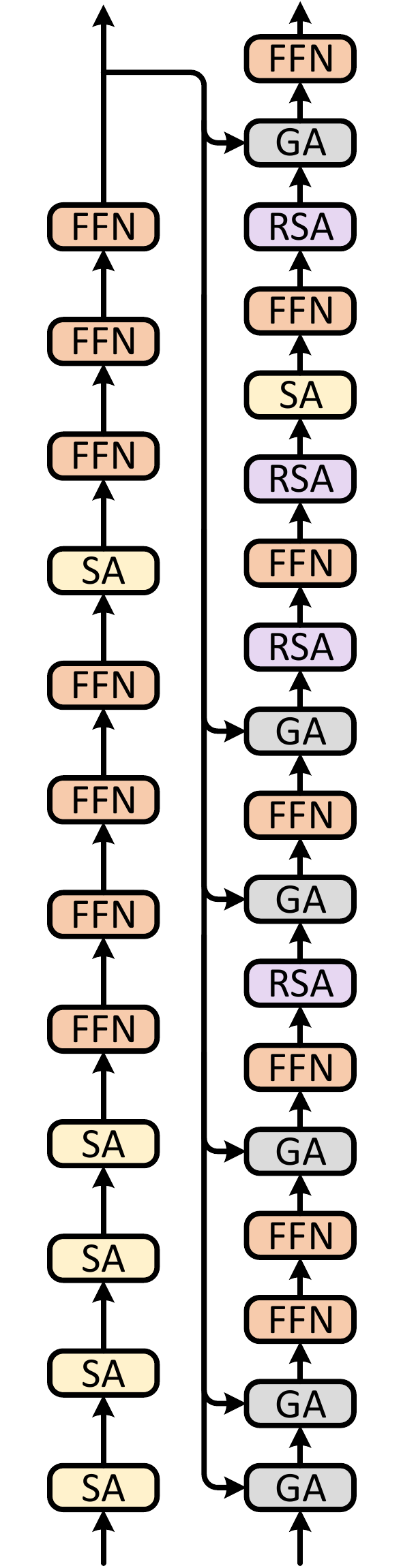}
        \caption{\makecell{VQA Task\\(VQA-v2)}}\label{fig:vqa_arch}
    \end{subfigure}
    \begin{subfigure}[h]{0.325\columnwidth}
        \includegraphics[width=\linewidth]{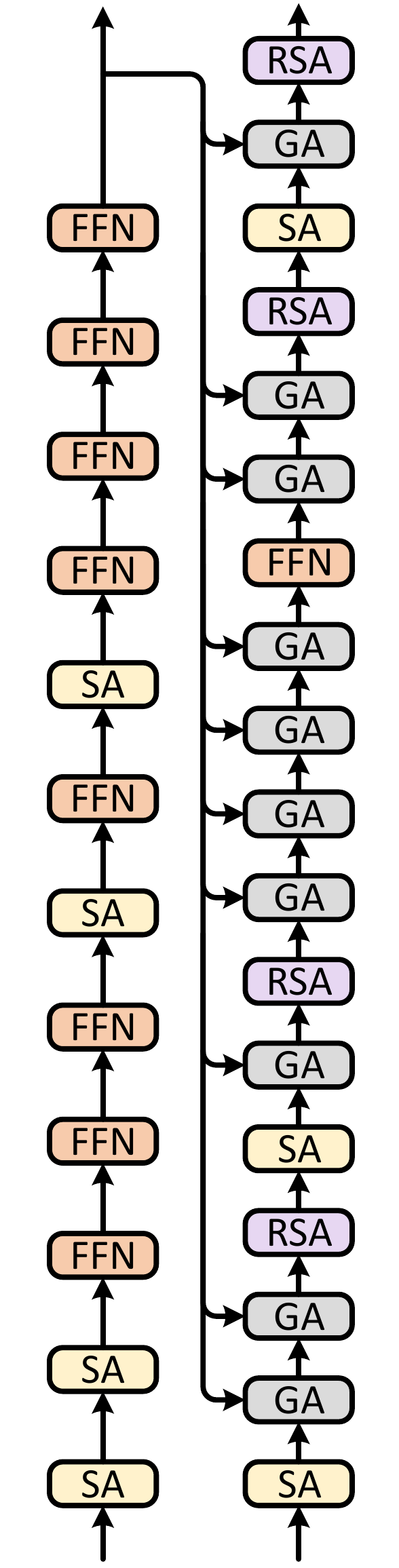}
        \caption{\makecell{ITM Task\\(Flickr30K)}}\label{fig:itm_arch}
    \end{subfigure}
    \begin{subfigure}[h]{0.325\columnwidth}
        \includegraphics[width=\linewidth]{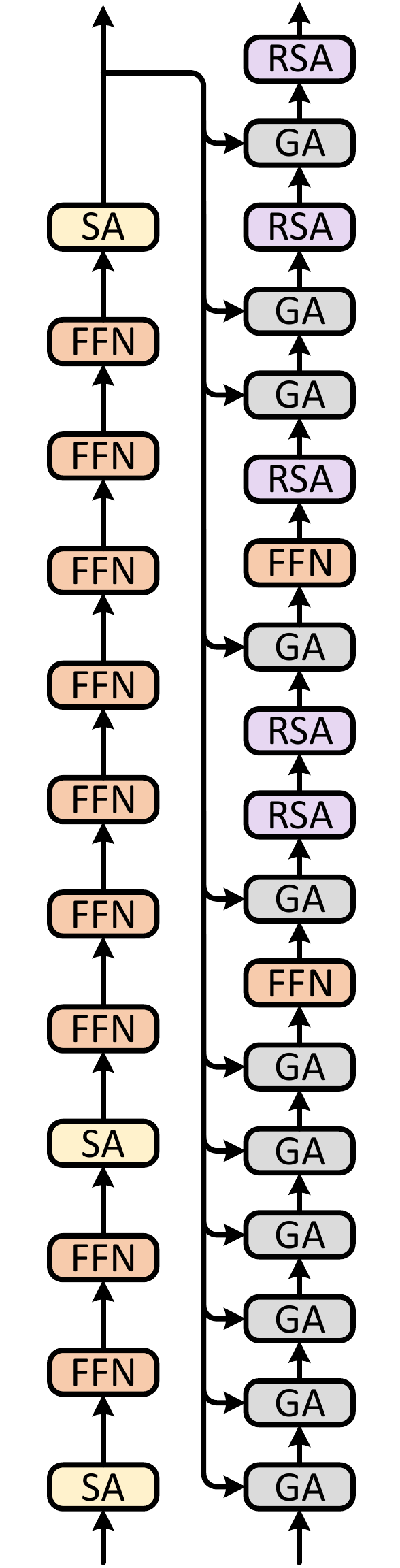}
        \caption{\makecell{VG Task\\(RefCOCO)}}\label{fig:vg_arch}
    \end{subfigure}
    \caption{The optimal MMnasNet backbones searched for different tasks (over specific datasets).}
    \label{fig:mmnasnet_arch}
\end{figure}


\setcounter{table}{4}
\begin{table*}
\centering
\small
\caption{Accuracies (with IoU$>$0.5) on RefCOCO, RefCOCO+ and RefCOCOg to compare with the state-of-the-art methods. All methods use \emph{detected objects} to extract visual features. COCO \cite{lin2014microsoft} and Genome \cite{krishna2017visual} denote two datasets for training the object detectors. SSD \cite{liu2016ssd}, FRCN \cite{ren2015faster} and MRCN \cite{he2017mask} denote the used detectors with VGG-16 \cite{simonyan2014very} or ResNet-101 \cite{he2015deep} backbones.
}\label{table:refcoco}
\begin{tabular}{l|ccc|ccc|ccc|cc}
\toprule
\multirow{3}{*}{Method} &\multicolumn{3}{c|}{Object Detector}& \multicolumn{3}{c|}{RefCOCO} & \multicolumn{3}{c}{RefCOCO+} & \multicolumn{2}{|c}{RefCOCOg}\\
\cmidrule{2-12}
&Dataset&Model&Backbone&  TestA & TestB & Val & TestA & TestB & Val & Test & Val \\
\midrule
CMN \cite{hu2017modeling} &COCO& FRCN &VGG-16 & 71.0 & 65.8 & - & 54.3 & 47.8 & - & - & -\\
VC \cite{zhang2018grounding} &COCO& FRCN &VGG-16 & 73.3 & 67.4 & - & 58.4 & 53.2 & - & - & -\\
\textbf{Spe.}+Lis.+Rein.+MMI \cite{yu2017joint} &COCO& SSD &VGG-16 & 73.7 & 65.0 & 69.5 & 60.7 & 48.8 & 55.7 & 59.6 & 60.2\\
Spe.+\textbf{Lis.}+Rein.+MMI \cite{yu2017joint} &COCO& SSD &VGG-16 & 73.1 & 64.9 & 69.0 & 60.0 & 49.6 & 54.9 & 59.2 & 59.3\\
MAttNet \cite{yu2018mattnet} &COCO& MRCN &ResNet-101 & 81.1 & 70.0 & 76.7 & 71.6 & 56.0 & 65.3 & 67.3 & 66.6 \\
DDPN \cite{yu2018rethinking} &Genome& FRCN &ResNet-101 & 80.1 & 72.4 & 76.8 & 70.5 & 54.1 & 64.8 & 67.0 & 66.7 \\
\midrule
MMnasNet (ours) & COCO & MRCN &ResNet-101 & 82.5 & 78.4 & 81.5 & 70.9 & 62.3 & 69.8 & 72.7 & 73.1\\
MMnasNet (ours) & Genome & FRCN &ResNet-101 & \textbf{87.4} & {77.7} & \textbf{84.2} & \textbf{81.0} & \textbf{65.2} & \textbf{74.7} & \textbf{75.7} & \textbf{74.7} \\
\bottomrule
\end{tabular}
\end{table*}

\setcounter{table}{2}
\begin{table}
\centering
\small
\caption{Accuracies on the \emph{test-dev} and \emph{test-std} splits of VQA-v2. All methods use the same visual features \cite{anderson2017up-down} and are trained on the \emph{train+val+vg} splits, where \emph{vg} denotes the augmented dataset from Visual Genome.}
\label{table:vqav2}
\begin{tabular}{l|cccc|c}
\toprule
\multirow{3}{*}{Method} & \multicolumn{4}{c|}{Test-Dev} &Test-Std \\
\cmidrule{2-6}
 & All & Y/N & Num & Other & All \\
\midrule
Bottom-Up \cite{teney2018tips} &65.32& 81.82& 44.21& 56.05 &65.67\\
MFH+CoAtt \cite{yu2018beyond}& 68.76 &84.27 &49.56 &59.89& -\\
BAN-8 \cite{kim2018bilinear}& 69.52 &85.31& 50.93& 60.26&-\\
BAN-8 (+G+C) \cite{kim2018bilinear} & 70.04& 85.42& {54.04}&60.52&70.35 \\
DFAF-8 \cite{gao2019dynamic} & 70.22& 86.09& 53.32 &60.49 & 70.34 \\
MCAN-6 \cite{yu2019mcan} & 70.63 & {86.82} & 53.26 & 60.72 & 70.90 \\
\midrule
MMnasNet (ours) & \textbf{71.24} & \textbf{87.27} &\textbf{55.68} &\textbf{61.05} & \textbf{71.46}\\
\bottomrule
\end{tabular}
\end{table}

\subsection{Main Results}
Taking the ablation studies into account, we compare the best-performing MMnasNet models (with $M$=12 and $N$=18) to the state-of-the-art approaches on five benchmark datasets. Figure \ref{fig:mmnasnet_arch} illustrates the optimal MMnasNet backbones searched for different tasks (over specific datasets). This verifies our hypothesis that the optimal architectures for different tasks may vary prominently. Note that we do not compare MMnasNet to the multimodal-BERT approaches (\emph{e.g.}, LXMRET \cite{tan2019lxmert} or UNITER \cite{chen2019uniter}), since they introduce additional training datasets for model pre-training thus may lead to unfair comparison.

In Table \ref{table:vqav2}, we compare MMnasNets to the state-of-the-art methods on VQA-v2. The demonstrated results show that: 1) compared with existing state-of-the-art approaches, MMnasNet outperforms them by a clear margin on all answer types; and 2) MMnasNet significantly outperforms existing approaches on the object counting performance (\emph{i.e.}, the number type), which owes to the effectiveness of the RSA operation we introduce.

\setcounter{table}{3}
\begin{table}
\centering
\small
\caption{Recall@\{1, 5, 10\} on Flickr30K to compare with the state-of-the-art methods.}
\label{table:flickr30k}
\begin{tabular}{l|ccc|ccc}
\toprule
\multirow{3}{*}{Method} & \multicolumn{3}{c|}{Text Retrieval} &\multicolumn{3}{c}{Image Retrieval} \\
\cmidrule{2-7}
 & R@1 &  R@5 &  R@10 & R@1 &  R@5 &  R@10 \\
\midrule
VSE++ \cite{faghri2017vse} & 52.9 &80.5 &87.2 & 39.6 &70.1 &79.5 \\
DAN \cite{nam2017dual} & 55.0 &81.8& 89.0 &39.4& 69.2& 79.1\\
DPC \cite{zheng2017dual} & 55.6 &81.9& 89.5 &39.1 &69.2 &80.9\\
SCO \cite{huang2018learning} &55.5 &82.0& 89.3 &41.1& 70.5& 80.1\\
SCAN$_\mathrm{t\rightarrow i}$ \cite{lee2018stacked} & 61.8& 87.5& 93.7& 45.8 &74.4& 83.0\\
SCAN$_\mathrm{i\rightarrow t}$ \cite{lee2018stacked} & 67.7 &88.9& 94.0& 44.0& 74.2 &82.6\\
CAMP \cite{wang2019camp} & 68.1 &89.7& 95.2& 51.5& 77.1& 85.3\\
\midrule
MMnasNet (ours)& \textbf{78.3} & \textbf{94.6} & \textbf{97.4} & \textbf{60.7} & \textbf{85.1} & \textbf{90.5} \\
\bottomrule
\end{tabular}
\end{table}

Table \ref{table:flickr30k} contains the image-text matching results on Flickr30K. Similar to most existing works \cite{lee2018stacked, wang2019camp}, we report the matching results in terms of Recall@$K$, where $K$ denotes the top-$K$ results retrieved from a database and ranges within $\{1,5,10\}$.  The cross-modal matching results from two directions (\emph{i.e.}, the text retrieval and image retrieval) are demonstrated in Table \ref{table:flickr30k} to compare with the state-of-the-art approaches. From the results, we can see that MMnasNet significantly outperforms existing state-of-the-art methods in terms of all evaluation metrics.

In Table \ref{table:refcoco}, we report the comparative results on RefCOCO, RefCOCO+, and RefCOCOg, respectively. We use the commonly used accuracy metric \cite{yu2018mattnet}, where a prediction is considered to be correct if the predicted bounding box overlaps with the ground-truth of IoU $>$0.5. With the standard visual features (\emph{i.e.}, the MRCN model pre-trained on COCO), MMnasNet reports a remarkable improvement over MAttNet on all the three datasets. Be equipped with the powerful visual features (\emph{i.e.}, the FRCN model pre-trained on Visual Genome), MMnasNet obtains further improvement and delivers the new state-of-the-art performance across all datasets.

\section{Conclusion}
In this paper, we present a generalized deep multimodal neural architecture search (MMnas) framework for various multimodal learning tasks. Different from the existing approaches that design hand-crafted and task-specific architectures to address only a single task, MMnas can be generalized to automatically learn the optimal architectures of different tasks. To achieve this,
we construct a unified encoder-decoder backbone with each encoder/decoder block corresponding to an operation searched from a candidate set of predefined operations. On top of the unified backbone, we attach task-specific heads to deal with different tasks. The optimal architecture for each task is learned by an efficient neural architecture search (NAS) algorithm to obtain task-specific MMnasNet. Extensive experiments are conducted on the VQA, visual grounding, and image-text matching tasks to show the generalizability and effectiveness of the proposed MMnas framework. Comprehensive results from five benchmark datasets validate the superiority of MMnasNet over existing state-of-the-art methods.

Different from existing multimodal-BERT approaches that use large-scale multimodal pre-training, we introduce an alternative way to address the generalized multimodal learning problem via a NAS framework. We hope our work may serve as a solid baseline to inspire future research on multimodal learning.

\begin{acks}
This work was supported in part by the National Key R\&D Program of China under Grant 2018AAA0100603, and in part by National Natural Science Foundation of China under Grant 61702143, Grant 61836002, Grant 61725203 and Grant 61732008.
\end{acks}
\bibliographystyle{ACM-Reference-Format}
\bibliography{mmfp0480.bbl}


\end{document}